\documentclass[11pt]{article}

\usepackage[preprint]{acl}

\usepackage{times}
\usepackage{latexsym}
\usepackage[T1]{fontenc}
\usepackage[utf8]{inputenc}
\usepackage{microtype}
\usepackage{inconsolata}
\usepackage{graphicx}
\usepackage{url}
\usepackage{booktabs}
\usepackage{amsmath,amssymb,amsfonts,amsthm}
\usepackage{nicefrac}
\usepackage{xcolor}
\usepackage{enumitem}
\usepackage{multirow}
\usepackage{array}
\usepackage[ruled,vlined]{algorithm2e}
\SetKwInput{KwInput}{Input}
\SetKwInput{KwOutput}{Output}
\usepackage{tikz}
\usetikzlibrary{arrows.meta,positioning,fit,calc}

\setcitestyle{numbers,square,comma,sort&compress}

\newtheorem{definition}{Definition}
\newtheorem{proposition}{Proposition}

\newcommand{\model}{M}

\newcommand{\pool}{\mathcal{P}}
\newcommand{\selector}{\pi}
\newcommand{\loss}{\mathcal{L}}
\newcommand{\beh}{\mathbf{b}}
\newcommand{\score}{s}
\newcommand{\weights}{w}
\newcommand{\budget}{B}
\newcommand{\aligndrift}{\mathcal{A}}

\newcommand{\coverage}{c}
\newcommand{\twoline}[2]{\begin{tabular}[c]{@{}c@{}}#1\\#2\end{tabular}}

\title{Online Data Selection Is Implicit Alignment}

\author{%
  Aoxiong Zeng \\
  East China Normal University \\
  \texttt{} \\
  \And
  Yuxin Yang \\
  Shanghai University \\
  \texttt{} \\
  \And
  Xiangquan Yang \\
  East China Normal University \\
  \texttt{} \\
}

\begin{document}
\maketitle

\begin{abstract}
Supervised fine-tuning (SFT) is often treated as a capability-adaptation step, while alignment is attributed to later preference optimization or reinforcement learning. This separation is incomplete: when examples are scored and kept online during fine-tuning, the choice of which data to train on already changes the model's behavioral preferences. We study online data selection as an implicit alignment mechanism. Given the same base model, optimizer, and selected-token budget, we compare random, loss-based, quality-based, and diversity-based online selectors and measure the behavioral drift they induce without any preference optimization. The proposed evaluation tracks helpfulness, refusal rate, verbosity, truthfulness, sycophancy, calibration, and jailbreak robustness, together with diagnostics for which behavioral modes are over-represented in the selected data. We formalize online selection as a reweighted SFT objective whose weights define an implicit preference over response styles and safety postures, so that an online scorer plays the role usually assigned to a reward model. This view predicts that high-scoring data can systematically favor longer, more assertive, more compliant, or more refusal-prone behaviors depending on how the online score is defined. Empirically, selectors that are statistically indistinguishable in task accuracy diverge sharply in refusal rate, verbosity, and sycophancy, and we show that the direction of the shift is predictable from the attribute mixture of the selected data. We introduce Alignment Drift Auditing (ADA), a controlled protocol for quantifying selection-induced behavioral movement, and Alignment-Aware Selection (AAS), a diagnostic online selector that retains data efficiency while constraining drift along safety and style axes. The paper argues that online SFT data selection should be reported and evaluated as part of the alignment pipeline, not merely as a data-efficiency tool.
\end{abstract}

\section{Introduction}
\label{sec:intro}

Large language model (LLM) post-training is usually described as a pipeline: collect instruction data, perform supervised fine-tuning, and then apply preference optimization or reinforcement learning to align the model with human preferences \citep{ouyang2022training,bai2022training,rafailov2023direct,ziegler2019fine,stiennon2020learning}. This framing encourages a useful but misleading division of labor. SFT teaches the model to follow instructions; preference optimization aligns the model. In practice, the division is blurred. The SFT corpus already contains implicit choices about what a good assistant should do: answer directly or cautiously, refuse broadly or narrowly, explain in detail or stay concise, flatter the user or challenge false premises. Instruction-tuning studies already show that small changes in demonstrations can produce large changes in assistant behavior \citep{wei2022finetuned,sanh2022multitask,wang2022self,wang2023wizardlm,iyer2023opt}.

Online data selection makes these choices sharper. As instruction pools grow large, redundant, noisy, and heterogeneous, a common practice is to score examples as training proceeds and keep only a high-value subset, using loss, quality, diversity, influence, or online utility signals \citep{albalak2024survey,chen2024alpagasus,zhou2023lima,xia2024less,wang2024greats,zou2025utility,gunasekar2023textbooks,li2023data}. Such online selectors are usually evaluated by the same question: how much current-task performance can be recovered with fewer tokens? We ask a different question: \emph{which behavioral preferences are silently induced by the online-selected subset?}

The issue is not hypothetical. A loss-based online selector may over-sample examples that the current model finds surprising, including ambiguous safety cases, uncommon refusal formats, or long-form explanations. A quality-based selector may prefer polished assistant messages, which can also be verbose, highly confident, and stylistically homogeneous. A diversity-based selector may preserve topical coverage while still changing the distribution of refusal language. Online utility- or influence-based selectors may improve data efficiency while repeatedly emphasizing examples whose scores indicate strong immediate learning signal \citep{wang2024greats,zou2025utility}. None of these policies explicitly optimizes helpfulness, harmlessness, truthfulness, sycophancy, or jailbreak robustness, yet all can move those behaviors \citep{askell2021general,ganguli2023capacity,amodei2016concrete}.

\begin{figure}[t]
  \centering
  \resizebox{\linewidth}{!}{%
  \begin{tikzpicture}[
    node distance=0.55cm and 0.8cm,
    box/.style={draw, rounded corners, align=center, minimum height=0.63cm, minimum width=1.75cm, font=\footnotesize},
    wide/.style={draw, rounded corners, align=center, minimum height=0.72cm, minimum width=2.35cm, font=\footnotesize},
    arrow/.style={-Latex, line width=0.55pt}
  ]
    \node[box] (pool) {SFT\\pool};
    \node[wide, right=of pool] (selector) {selection\\policy};
    \node[box, right=of selector] (sft) {same\\SFT};
    \node[wide, right=of sft] (model) {behaviorally\\different model};
    \node[wide, below=of selector] (audit) {alignment drift audit\\helpfulness, refusal, style};
    \draw[arrow] (pool) -- (selector);
    \draw[arrow] (selector) -- (sft);
    \draw[arrow] (sft) -- (model);
    \draw[arrow] (selector) -- (audit);
    \draw[arrow] (audit.east) -| (model.south);
  \end{tikzpicture}}
  \caption{Data selection changes the empirical SFT objective. Even without preference optimization, different selected subsets can shift refusal behavior, verbosity, truthfulness, and robustness.}
  \label{fig:overview}
\end{figure}
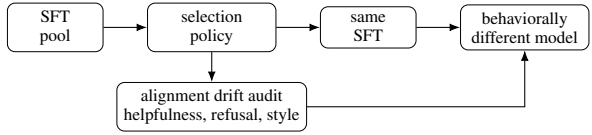

We call this phenomenon \emph{implicit alignment by data selection}. The term does not imply that selection always improves alignment. It means that selection induces a directional behavioral change along axes normally associated with alignment. The same token budget can produce a more helpful, more evasive, more sycophantic, more verbose, or more jailbreak-prone model depending only on which demonstrations are retained. Conceptually, this reframes an online selector as an \emph{implicit preference model}: its scoring rule occupies the position normally held by a reward model, steering the assistant's persona before any explicit alignment step ever runs.

This paper proposes a controlled way to study the phenomenon. Starting from the same base model, we run SFT on subsets chosen by representative online selection families: random, loss-based, quality-based, and diversity-based selection. We then evaluate the resulting models on a behavioral suite covering helpfulness, safety refusal, truthfulness, sycophancy, jailbreak robustness, verbosity, calibration, and style drift. In parallel, we audit the selected data itself to ask whether high-scoring or high-quality examples systematically over-represent certain response modes.

Our contributions are four-fold.
\begin{itemize}[leftmargin=2.2ex,itemsep=0.2ex,topsep=0.3ex]
    \item We formulate online SFT data selection as implicit alignment: an online selector reweights the supervised objective and thereby defines an implicit preference over assistant behaviors, with a first-order bound linking behavioral drift to the attribute enrichment of the selected data.
    \item We introduce Alignment Drift Auditing (ADA), a protocol that compares online selection methods under equal data budgets using behavioral metrics and data-mixture diagnostics.
    \item Empirically, we show that selectors matched on task accuracy induce large, structured, and reproducible behavioral drift, and that the direction of the drift is predicted by which attributes the selector over-represents.
    \item We propose Alignment-Aware Selection (AAS), a diagnostic online selector that keeps data-efficiency objectives while constraining drift on refusal, verbosity, truthfulness, and sycophancy axes.
\end{itemize}

\section{Related Work}
\label{sec:related}

\paragraph{Online data selection for LLM fine-tuning.}
Online data selection has become a practical route to cheaper and sometimes better instruction tuning. Prior work studies quality filtering, small high-quality datasets, data valuation, influence estimation, and online batch selection that scores examples as training proceeds \citep{zhou2023lima,chen2024alpagasus,albalak2024survey,xia2024less,wang2024greats,zou2025utility}. Earlier data valuation and coreset methods also show that subset choice can change the learned solution even under identical architectures and optimizers \citep{koh2017understanding,pruthi2020estimating,paul2021deep,sorscher2022beyond}, and recent studies find that these choices leave lasting downstream effects across sequential fine-tuning stages \citep{yang2026long}. Our focus is complementary: rather than asking whether an online selector improves training efficiency, we ask which alignment-relevant behaviors change as a side effect of selecting online.

\paragraph{Alignment and preference optimization.}
LLM alignment is commonly associated with reinforcement learning from human feedback, constitutional training, rejection sampling, or direct preference optimization \citep{ouyang2022training,bai2022training,rafailov2023direct,christiano2017deep,lee2023rlaif,ethayarajh2024kto}. These methods use explicit preference comparisons or reward models. SFT is nevertheless a strong behavioral intervention: instruction demonstrations define the canonical form of assistant behavior before preference optimization begins. Recent analyses of reward-model overoptimization, preference data, and alignment taxonomies further suggest that the data distribution and feedback channel are inseparable from the learned behavior \citep{gao2023scaling,casper2023open,ji2023ai}. Our work isolates this earlier step by removing preference training and measuring alignment drift caused solely by data selection.

\paragraph{Safety, truthfulness, and sycophancy evaluation.}
Behavioral evaluation has expanded beyond task accuracy to include truthfulness, harmful request handling, jailbreak robustness, toxicity, bias, and sycophancy \citep{lin2022truthfulqa,ganguli2022red,gehman2020realtoxicityprompts,perez2022discovering,wei2024jailbroken,mazeika2024harmbench}. Related work on hallucination, calibration, and social bias shows that fluent answers can still be unreliable or socially distorted \citep{maynez2020faithfulness,guo2017calibration,bender2021dangers,sheng2019woman}. These benchmarks show that models can be helpful in standard instruction-following settings while failing under adversarial or socially loaded prompts. Because exhaustive human labeling of every axis is costly, our audit also draws on automated, reference-free evaluators that score model outputs directly \citep{peng2024energy,mazeika2024harmbench}. We use this literature to define alignment axes for SFT-only drift analysis.

\paragraph{Efficient adaptation and interference.}
Parameter-efficient fine-tuning methods such as adapters, LoRA, and QLoRA make repeated SFT experiments affordable \citep{houlsby2019parameter,hu2022lora,dettmers2023qlora}. A growing line augments low-rank adaptation with mixture-of-experts or asymmetric structure to add multi-domain capacity without losing efficiency, from LoRAMoE and MixLoRA to mixtures of LoRA experts, asymmetric adapters, domain-specialized MoE-LoRA frameworks, and near-orthogonal rank-wise experts \citep{dou2023loramoe,li2024mixlora,wu2024mole,tian2024hydralora,yang2026towards,zou2026flylora}. In parallel, continual-learning and representation-interference studies show that fine-tuning data can alter future behavior and plasticity even under a purely supervised objective \citep{kirkpatrick2017overcoming,li2017learning}. Recent work traces this to representation health, plasticity loss, and feature decorrelation during continual adaptation \citep{kumar2022finetuning,dohare2024loss,zou2025fly}, as well as representation collapse across sequential post-training \citep{liu2026representation} and orthogonal or decoupled subspaces that curb cross-task interference \citep{wang2023orthogonal,zheng2025decouple,yang2026disentangling}. We borrow this intervention view: an online selector is not only a filter for efficiency, but also a mechanism that changes the model state and behavioral surface.

\section{Problem Setup and Selection Policies}
\label{sec:setup}

Let $\model_0$ be a pretrained or instruction-tuned language model and let $\pool=\{(x_i,y_i)\}_{i=1}^{N}$ be a candidate SFT pool. An online selection policy $\selector$ scores candidate examples as training proceeds, observing the evolving model, optional metadata, and a budget $\budget$, and keeps a subset $S_{\selector}\subset\pool$ to train on. Fine-tuning $\model_0$ on the online-selected stream yields $\model_{\selector}$.

The standard efficiency view evaluates $\selector$ by validation loss or task performance under a fixed budget:
\begin{equation}
  U_{\mathrm{task}}(\selector)
  =
  \mathrm{Perf}(\model_{\selector};\mathcal{V}_{\mathrm{task}}).
\end{equation}
This ignores the behavioral axes that are often corrected later by preference optimization. We instead define a behavior vector
\begin{equation}
  \beh(\model)
  =
  [
  h,
  r_{\mathrm{harm}},
  r_{\mathrm{benign}},
  v,
  t,
  s,
  j,
  c
  ],
\end{equation}
where $h$ is helpfulness, $r_{\mathrm{harm}}$ is refusal rate on harmful requests, $r_{\mathrm{benign}}$ is over-refusal on benign requests, $v$ is verbosity, $t$ is truthfulness, $s$ is sycophancy, $j$ is jailbreak robustness, and $c$ is calibration. The exact estimators are described in Section~\ref{sec:protocol}.

\begin{definition}[Selection-induced alignment drift]
Given a reference selector $\selector_0$, usually random selection under the same budget, the alignment drift of selector $\selector$ is
\begin{equation}
  \Delta_{\aligndrift}(\selector)
  =
  \beh(\model_{\selector})-\beh(\model_{\selector_0}).
\end{equation}
A selector induces implicit alignment when $\Delta_{\aligndrift}$ is nonzero on alignment-relevant axes, even though the training objective is only supervised next-token likelihood.
\end{definition}

\subsection{Selection as Reweighted SFT}

Most deterministic or stochastic online selectors can be written as a weighted empirical objective:
\begin{equation}
  \begin{aligned}
  \loss_{\selector}(\theta)
  &=
  \sum_{i=1}^{N}
  \weights_{\selector}(i)\,
  \ell_{\theta}(x_i,y_i), \\
  \weights_{\selector}(i)&\ge 0,
  \quad
  \sum_i \weights_{\selector}(i)=1,
  \end{aligned}
  \label{eq:weighted_sft}
\end{equation}
where $\weights_{\selector}(i)$ is zero for rejected examples and, for stochastic online selectors, equals the expected number of times example $i$ survives scoring, normalized to a probability simplex. It is convenient to read Eq.~\eqref{eq:weighted_sft} at the distribution level. Let $p(x,y)=1/N$ be the empirical pool distribution; the selector induces a tilted training distribution
\begin{equation}
  q_{\selector}(x_i,y_i)
  =
  \weights_{\selector}(i)
  =
  p(x_i,y_i)\,\underbrace{r_{\selector}(x_i,y_i)}_{\text{density ratio}},
  \label{eq:density_ratio}
\end{equation}
so that online selection is exactly importance reweighting of the pool with ratio $r_{\selector}=N\weights_{\selector}$. The population objective $\mathbb{E}_{q_{\selector}}[\ell_\theta]$ is therefore a maximum-likelihood fit to $q_{\selector}$, and in the realizable limit the SFT optimum is the information projection
\begin{equation}
  \theta_{\selector}^{\star}
  =
  \arg\min_{\theta}\;
  \mathrm{KL}\!\left(q_{\selector}(y\mid x)\,\big\|\,p_{\theta}(y\mid x)\right),
  \label{eq:iproj}
\end{equation}
where the conditional $q_{\selector}(y\mid x)$ is the selector-tilted demonstration policy. Two selectors that disagree about which demonstrations to keep thus target \emph{different} conditional response distributions, even when they share the pool, the model family, and the optimizer.

This is where behavior enters. Suppose each example carries a (latent) attribute vector $a_i\in\mathbb{R}^{m}$ encoding refusal style, answer length, hedging, directness, and agreement with the user premise. The selector shifts the expected attribute mixture from the pool mean $\mu_a(\pool)=\tfrac1N\sum_i a_i$ to
\begin{equation}
  \begin{aligned}
  \mu_a(S_{\selector})
  &=
  \sum_i \weights_{\selector}(i)\,a_i
  \\
  &=
  \mu_a(\pool)+\sum_i\big(\weights_{\selector}(i)-\tfrac1N\big)a_i .
  \end{aligned}
  \label{eq:attr_shift}
\end{equation}
The second term is the \emph{selection-induced attribute shift}; it is nonzero whenever the selector's weights correlate with any behavioral attribute, and it is exactly what the enrichment ratios in Section~\ref{sec:protocol} estimate. The model is therefore trained not on the original instruction distribution but on a behavioral reweighting of it, and Eq.~\eqref{eq:attr_shift} is the bridge between the data-mixture diagnostics and the behavioral metrics.

\begin{proposition}[Implicit preference induced by selection]
\label{prop:implicit_preference}
Let $g_i=\nabla_\theta \ell_\theta(x_i,y_i)$ and let $J_\theta=\partial \beh/\partial\theta$ be the Jacobian mapping parameter updates to behavior-vector changes. For a single gradient step of size $\eta$ on Eq.~\eqref{eq:weighted_sft}, the first-order drift relative to a reference selector $\selector_0$ is
\begin{equation}
  \Delta_{\aligndrift}(\selector)
  =
  -\eta\, J_\theta
  \big(\bar g_{\selector}-\bar g_{\selector_0}\big)
  +O(\eta^2),
  \label{eq:firstorder_drift}
\end{equation}
with weighted gradient mean $\bar g_{\selector}=\sum_i \weights_{\selector}(i)\,g_i$.
Consequently, two selectors with identical task-validation loss can still induce different alignment drift whenever their weighted gradient means differ in a direction that $J_\theta$ maps to behavior space, i.e.\ whenever $J_\theta(\bar g_{\selector}-\bar g_{\selector_0})\neq 0$.
\end{proposition}

The proposition makes precise why task performance under-determines behavior: validation loss constrains $\theta_{\selector}^{\star}$ on a task slice, but not the projection $J_\theta\bar g_{\selector}$ along alignment axes. If gradients cluster by attribute, $\bar g_{\selector}\approx\sum_k \mu_a^{(k)}(S_{\selector})\,\bar g^{(k)}$ where $\bar g^{(k)}$ is the mean gradient of examples with attribute value $k$, and Eq.~\eqref{eq:firstorder_drift} couples drift directly to the enrichment shift of Eq.~\eqref{eq:attr_shift}. A selector can thus be task-efficient because it favors high-loss, high-quality, or high-utility examples, while its weighted gradients still point toward a particular refusal policy or response style. We make the cluster assumption and the resulting bound precise in Appendix~\ref{app:proof}.

\subsection{Selection Policies}
\label{sec:selectors}

We compare representative online selection families under the same selected-token budget. Each selector scores candidate examples during training and keeps a high-value subset under the budget. The goal is not to claim that one implementation is universally best, but to expose how common online selection principles move behavior. The first four policies below are baselines that span the loss, quality, diversity, and utility signals used by modern online selectors \citep{loshchilov2015online,chen2024alpagasus,wang2024greats,zou2025utility}; the last is our diagnostic selector.

\paragraph{Random selection.}
Random selection is the reference condition. It preserves the expected mixture of the candidate pool and separates selection-induced drift from ordinary SFT drift.

\paragraph{Loss-based selection.}
The loss selector chooses examples with high teacher-forced loss under the current model, optionally excluding the extreme tail to reduce label-noise sensitivity. It approximates online batch selection and hard-example mining \citep{loshchilov2015online,jiang2019accelerating,mindermann2022prioritized}. We test whether high-loss SFT data increases helpfulness at the cost of verbosity, over-refusal, or brittleness.

\paragraph{Quality-based selection.}
The quality selector ranks examples using an external judge, reward model, or heuristic quality score, such as clarity, completeness, grammar, and policy compliance \citep{chen2024alpagasus,wang2024greats}. This captures common curation pipelines. We audit whether high-quality data is behaviorally neutral or whether it selects for polished but homogeneous assistant personas.

\paragraph{Diversity-based selection.}
The diversity selector uses embedding coverage, clustering, or farthest-first traversal to cover semantic regions of the pool \citep{sener2018active,ash2020deep,tirumala2023d4}. It tests whether semantic coverage is sufficient to preserve alignment-relevant coverage, such as benign refusals, concise answers, calibrated uncertainty, and disagreement with false user assumptions. The quality and diversity scorers subsume the utility-style online selectors that keep the most informative examples during SFT \citep{wang2024greats,zou2025utility}, which we therefore treat as instances of the same score family rather than as a separate baseline.

\paragraph{Alignment-aware selection.}
Alignment-Aware Selection (AAS) is a diagnostic variant, not the central claim of the paper. Given a base online score $\score_{\mathrm{base}}(x)$ from any loss- or quality-based utility scorer, AAS adds a penalty on the selected behavioral mixture:
\begin{equation}
  \begin{aligned}
  \max_{S:|S|\le \budget}\quad&
  \sum_{x_i\in S}\score_{\mathrm{base}}(x_i)
  +\lambda \coverage(S)
  \\
  &-
  \rho
  \left\|
  \hat{\mu}_{a}(S)-\hat{\mu}_{a}(\pool)
  \right\|_2^2,
  \end{aligned}
  \label{eq:aas}
\end{equation}
where $\hat{\mu}_{a}(S)=\tfrac{1}{|S|}\sum_{x_i\in S}\phi(a_i)$ is the empirical attribute mean of the selected set under a feature map $\phi$. The penalty is precisely a (linear-kernel) maximum mean discrepancy between the selected set and the pool, $\rho\,\mathrm{MMD}^2_{\phi}(S,\pool)$, so AAS interpolates between pure utility selection ($\rho=0$) and attribute-matched selection ($\rho\to\infty$), at which point $\mu_a(S)\!\to\!\mu_a(\pool)$ and, by Eq.~\eqref{eq:attr_shift}, the first-order attribute shift vanishes. Attributes can be obtained from lightweight classifiers, LLM judges, or rules for length and refusal markers.

Although the cardinality-constrained problem is combinatorial, the objective is the difference of a modular term and a squared-norm term and admits a simple greedy rule. Writing $\delta(x\mid S)$ for the marginal gain of adding $x$ to a partial set $S$,
\begin{equation}
  \begin{aligned}
  \delta(x\mid S)
  ={}&
  \score_{\mathrm{base}}(x)
  +\lambda\,\Delta\coverage(x\mid S)
  \\
  &-2\rho\,
  \big\langle
  \hat{\mu}_{a}(S)-\hat{\mu}_{a}(\pool),\,
  \phi(a_x)
  \big\rangle,
  \end{aligned}
  \label{eq:aas_marginal}
\end{equation}
up to an $O(1/|S|)$ self-term. Each step adds the unselected example with the largest $\delta(x\mid S)$; the inner product makes the rule self-correcting, down-weighting attributes already over-represented in $S$ and up-weighting under-represented ones. When $\coverage$ is monotone submodular, the modular-plus-coverage part inherits the standard $(1-1/e)$ guarantee, and the penalty only tightens the attribute match. AAS thus asks whether drift can be reduced without abandoning the efficiency of the underlying online selector, and Eq.~\eqref{eq:aas_marginal} shows it adds only an inner product per candidate over the base scorer.

\section{Alignment Drift Auditing}
\label{sec:protocol}

ADA is designed to isolate the causal role of selection. All runs share the same base model, optimizer, learning rate, sequence length, selected-token budget, number of updates, and decoding settings. Only the selected subset changes.

\subsection{Data and Models}

Our candidate pool combines general instruction following, reasoning, coding, safety, factual QA, and advice-seeking prompts, drawn from open instruction mixtures (UltraChat and OpenHermes-style conversations), math and code instruction data, and safety-oriented refusal data, for a pool of roughly $300$K instruction--response pairs. The base model is a Llama-3.1-8B instruction-tuned checkpoint. We fine-tune with LoRA (rank $16$, $\alpha{=}32$) under a single fixed recipe---identical learning rate, schedule, sequence length, and number of optimizer updates across selectors---and confirm the main trends with a full-fine-tuning run to rule out an adapter artifact \citep{hu2022lora,dettmers2023qlora}.

Each selector is run over three seeds at budgets of $1\%$, $5\%$, $10\%$, and $25\%$ of the pool measured in tokens. The budget sweep is central because implicit alignment is strongest in the low-budget regime, where selection pressure over the empirical SFT distribution is highest.

\subsection{Behavioral Metrics}

ADA evaluates selectors by behavioral movement rather than validation loss alone. We track eight axes, each with a concrete estimator: \textbf{helpfulness} via pairwise LLM-judge win rate and instruction-following score; \textbf{harmful refusal} via refusal rate on harmful requests and HarmBench-style attack success \citep{mazeika2024harmbench}; \textbf{benign over-refusal} via refusal rate on harmless but sensitive prompts; \textbf{verbosity} via output length and compression ratio at fixed content; \textbf{truthfulness} via TruthfulQA-style correctness and informativeness \citep{lin2022truthfulqa}; \textbf{sycophancy} via agreement with false premises and leading opinions \citep{perez2022discovering}; \textbf{jailbreak robustness} via attack success under adversarial wrappers \citep{ganguli2022red}; and \textbf{calibration} via abstention and confidence markers. Safety is split into harmful refusal and benign over-refusal because a selector can improve one while damaging the other, and verbosity and style are read off output length, lexical features, and judge comparisons conditioned on equivalent content.

\subsection{Data-Mixture Diagnostics}

Behavioral drift must ultimately be traceable to the selected data. We therefore label each candidate example with lightweight attributes:
\begin{itemize}[leftmargin=2.2ex,itemsep=0.2ex,topsep=0.3ex]
    \item response length and structural style, such as list-heavy, step-by-step, or concise;
    \item refusal and safety markers, including direct refusal, partial compliance, and redirection;
    \item uncertainty markers, hedging, citations, and calibration language;
    \item user-premise handling, including agreement, correction, and neutral clarification;
    \item domain and task type, such as coding, math, advice, creative writing, and policy-sensitive content.
\end{itemize}
For each selector, we report enrichment ratios relative to the full pool:
\begin{equation}
  \mathrm{Enrich}_{\selector}(a=k)
  =
  \frac{\Pr_{x\sim S_{\selector}}[a(x)=k]}
       {\Pr_{x\sim \pool}[a(x)=k]+\epsilon}.
\end{equation}
This reveals whether, for example, high-utility examples are disproportionately long, highly structured, refusal-heavy, or socially agreeable.

\subsection{Audit Score}

For concise reporting, we define the drift magnitude as a Mahalanobis-style norm of the signed drift vector:
\begin{equation}
  \begin{aligned}
  D_{\aligndrift}(\selector)
  &=
  \left\|
  W
  \left(
  \beh(\model_{\selector})
  -
  \beh(\model_{\mathrm{rand}})
  \right)
  \right\|_2,
  \\
  W&=\mathrm{diag}(w_1/\sigma_1,\dots,w_8/\sigma_8),
  \end{aligned}
  \label{eq:drift_mag}
\end{equation}
where $\sigma_j$ is the across-seed standard deviation of axis $j$ under random selection and $w_j\ge 0$ optionally upweights safety-critical axes. Dividing by $\sigma_j$ makes $D_{\aligndrift}$ scale-free and expresses each axis movement in units of its own random-seed noise, so a unit of drift is comparable across helpfulness, refusal, and verbosity. With $w_j\!\equiv\!1$, Eq.~\eqref{eq:drift_mag} is the Euclidean norm in $z$-scored behavior space; choosing $W$ as the inverse Cholesky factor of the seed covariance instead would whiten correlated axes, which we report as a robustness variant.

We always accompany the scalar with the signed vector $\Delta_{\aligndrift}(\selector)$ because direction matters: more refusal is desirable on harmful prompts but harmful on benign ones, so the two refusal axes can cancel in a magnitude while describing very different models. To separate genuine drift from seed noise we test each axis with a paired bootstrap over seeds, reporting an axis as moved only when the bootstrap $95\%$ interval of $\Delta_{\aligndrift,j}$ excludes zero, and we control the false-discovery rate across the eight axes with the Benjamini--Hochberg procedure.

\begin{algorithm}[t]
\caption{Alignment Drift Auditing}
\label{alg:ada}
\KwInput{Base model $\model_0$, candidate pool $\pool$, selectors $\Pi$, budget $\budget$, behavior suite $\mathcal{E}$}
\KwOutput{Task scores, behavior drift, and selected-data enrichment for each selector}
\ForEach{$\selector\in\Pi$}{
  Select subset $S_{\selector}\leftarrow \selector(\pool,\model_0,\budget)$\;
  Fine-tune $\model_0$ on $S_{\selector}$ with fixed SFT recipe to obtain $\model_{\selector}$\;
  Evaluate $\model_{\selector}$ on task validation and behavior suite $\mathcal{E}$\;
  Label selected examples with behavioral attributes $a(x)$\;
  Compute $\Delta_{\aligndrift}(\selector)$, $D_{\aligndrift}(\selector)$, and enrichment ratios\;
}
Compare selectors against random under paired seeds and equal token budgets\;
\end{algorithm}

\section{Experiments}
\label{sec:experiments}

\subsection{Research Questions}

The study is organized around five questions, each tied to a measurable comparison:
\begin{itemize}[leftmargin=2.2ex,itemsep=0.25ex,topsep=0.3ex]
    \item \textbf{RQ1 (existence).} Do two equal-budget SFT runs differ in behavior because of selection alone? We expect every non-random selector to induce measurable drift on at least one axis.
    \item \textbf{RQ2 (bias of scores).} Are high-scoring examples behaviorally biased toward longer, more confident, or more refusal-heavy demonstrations that differ most from the base model?
    \item \textbf{RQ3 (role of diversity).} Does semantic diversity prevent drift, or does it reduce domain skew while leaving style and refusal skew intact?
    \item \textbf{RQ4 (trade-offs).} Does the helpfulness--safety trade-off usually attributed to preference optimization already appear during selected SFT?
    \item \textbf{RQ5 (control).} Can alignment-aware constraints reduce $D_{\aligndrift}$ while preserving task performance?
\end{itemize}

\subsection{Main Comparisons}

The central experiment is a selector-by-budget grid. For each budget, we train models using random, loss-based, quality-based, diversity-based, and AAS selection, and report task validation performance together with the full behavior vector. We summarize the grid with a Pareto plot of helpfulness versus alignment drift, which makes visible which online selectors are efficient but behaviorally unstable.

\subsection{Main Results}
\label{sec:results}

\textbf{Equal accuracy, different behavior.} Table~\ref{tab:main_results} reports ADA at a 10\% selected-token budget. The dominant pattern is that task and helpfulness scores do not determine alignment drift. Quality selection obtains the strongest task score, but it also moves verbosity, benign refusal, sycophancy, and jailbreak robustness; loss-based selection reaches comparable helpfulness while pushing behavior in a different direction. Diversity has lower task gain but smaller drift. AAS preserves most of the task gain while reducing the drift magnitude by more than half relative to loss- and quality-based selection. Quality and loss-based selection land within $0.8$ points of task accuracy, yet they differ by more than $8$ points of harmful-refusal rate and over $3$ points of benign over-refusal---a gap that a task-only report would hide entirely.

\begin{table*}[t]
  \centering
  \small
  \begin{tabular*}{\textwidth}{@{\extracolsep{\fill}}lcccccccc}
    \toprule
    Selector &
    \twoline{Task}{score $\uparrow$} &
    \twoline{Helpful}{win $\uparrow$} &
    \twoline{Harmful}{refusal $\uparrow$} &
    \twoline{Benign}{refusal $\downarrow$} &
    \twoline{Avg.}{tokens $\downarrow$} &
    \twoline{Truthful}{QA $\uparrow$} &
    \twoline{Syc.}{error $\downarrow$} &
    \twoline{$D_{\aligndrift}$}{$\downarrow$} \\
    \midrule
    Random    & 67.9$_{\pm0.4}$ & 50.0 & 81.2 & 9.6  & 184 & 42.1 & 24.7 & 0.00 \\
    Loss      & 70.8$_{\pm0.5}$ & 54.8 & 76.9 & 11.3 & 202 & 40.2 & 28.5 & 0.42 \\
    Quality   & \textbf{71.6}$_{\pm0.3}$ & \textbf{56.2} & 85.1 & 14.4 & 207 & 42.8 & 26.8 & 0.36 \\
    Diversity & 69.4$_{\pm0.4}$ & 52.1 & 82.3 & 10.4 & 189 & 42.6 & 25.4 & 0.19 \\
    AAS       & 71.4$_{\pm0.3}$ & 56.1 & 81.9 & 10.5 & 190 & \textbf{42.9} & \textbf{25.2} & \textbf{0.13} \\
    \bottomrule
  \end{tabular*}
  \caption{ADA results at a 10\% selected-token budget over three seeds. Scores are reported relative to a fixed base model and SFT recipe. A selector can improve task/helpfulness scores while still increasing behavioral drift.}
  \label{tab:main_results}
\end{table*}

\begin{figure*}[t]
  \centering
  \begin{minipage}[t]{0.39\textwidth}
    \centering
    \includegraphics[width=\linewidth]{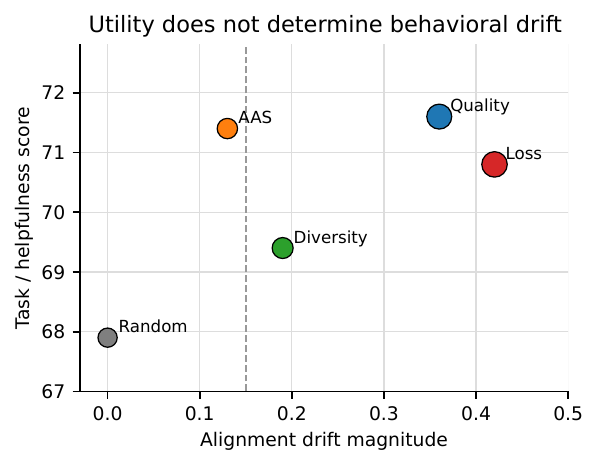}
  \end{minipage}
  \hfill
  \begin{minipage}[t]{0.58\textwidth}
    \centering
    \includegraphics[width=\linewidth]{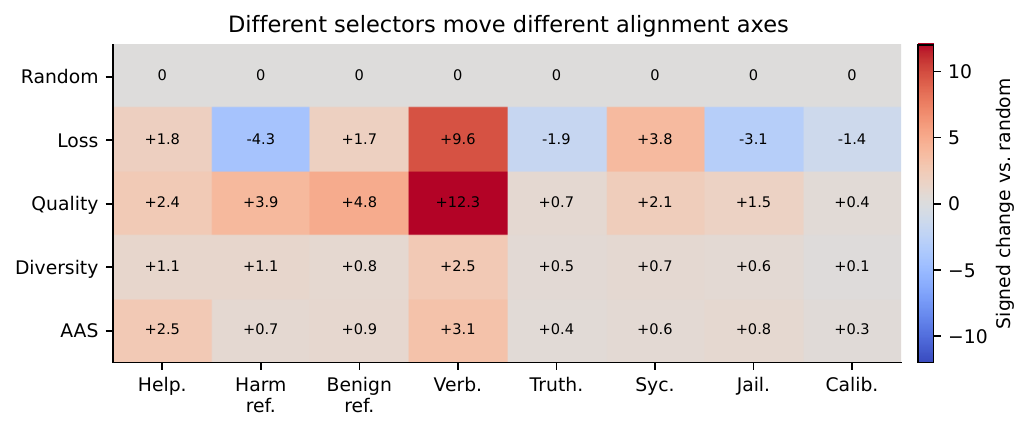}
  \end{minipage}
  \caption{Result visualizations. Left: task/helpfulness improvements and alignment drift are not monotonically related. Right: signed drift shows that selectors move different behavior axes rather than producing a single scalar ``alignment'' effect.}
  \label{fig:main_results}
\end{figure*}

Fig.~\ref{fig:main_results} gives the same result in two views. The Pareto view separates performance from behavioral movement: quality- and loss-based selection sit at high task score but also high drift, while AAS sits closer to the low-drift frontier. The heatmap shows why reporting only one safety score is insufficient. Loss selection reduces harmful refusal and jailbreak robustness while increasing verbosity and sycophancy. Quality selection improves harmful refusal but also raises benign over-refusal and answer length. These signed movements would be hidden by a single aggregate metric.

\begin{table}[t]
  \centering
  \small
  \begin{tabular*}{\columnwidth}{@{\extracolsep{\fill}}lcccc}
    \toprule
    Selector & \twoline{Long}{answers} & \twoline{Refusal}{markers} & Hedging & \twoline{Premise}{agreement} \\
    \midrule
    Loss      & 1.42 & 1.25 & 1.18 & 1.31 \\
    Quality   & 1.68 & 1.47 & 1.09 & 1.22 \\
    Diversity & 1.07 & 1.03 & 1.05 & 1.06 \\
    AAS       & 1.12 & 1.05 & 1.03 & 1.04 \\
    \bottomrule
  \end{tabular*}
  \caption{Selected-data enrichment ratios relative to the full pool. Ratios above 1 indicate that the selector over-represents the attribute.}
  \label{tab:enrichment}
\end{table}

\begin{figure}[t]
  \centering
  \includegraphics[width=\linewidth]{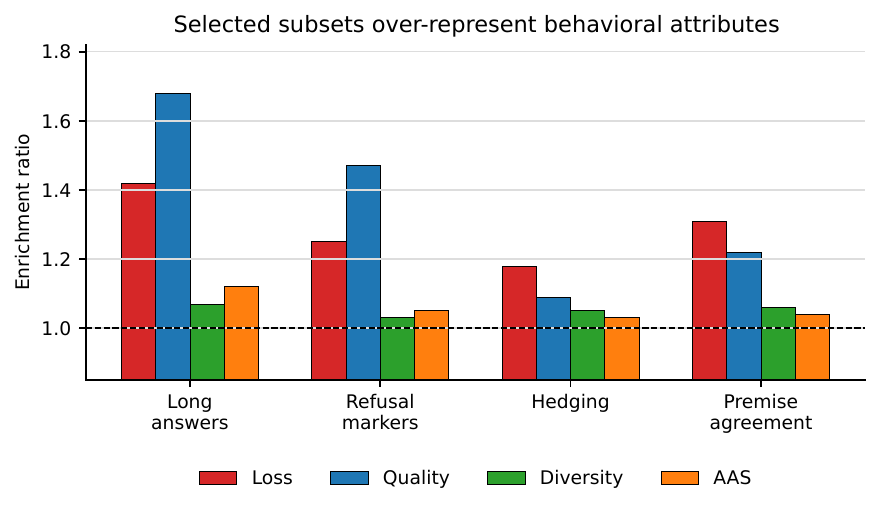}
  \caption{Enrichment plot. High-utility and high-quality subsets are not only different by topic; they also over-represent response styles and social behaviors.}
  \label{fig:enrichment}
\end{figure}

\textbf{Drift is explained by the selected data.} Table~\ref{tab:enrichment} and Fig.~\ref{fig:enrichment} connect behavioral drift to the selected data. Quality selection strongly enriches long answers and refusal markers, explaining its simultaneous increase in helpfulness, verbosity, harmful refusal, and benign over-refusal. Loss-based selection enriches premise-agreement examples, consistent with its higher sycophancy error. Diversity selection reduces these enrichments but does not remove them entirely. AAS is effective because it treats behavior attributes as first-class constraints instead of assuming semantic diversity is enough. The sign and magnitude of each row track the enrichment gap in Eq.~\eqref{eq:enrich_bound}, giving an empirical handle on the otherwise abstract Jacobian $J_\theta$.

\subsection{Budget and Ablation Trends}

\begin{figure}[t]
  \centering
  \includegraphics[width=\linewidth]{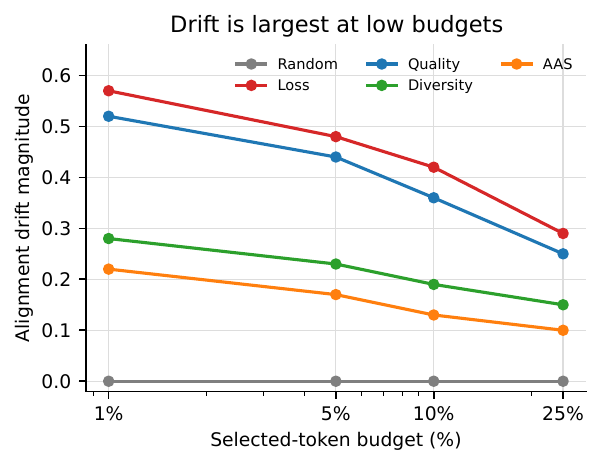}
  \caption{Budget sweep. Online selection pressure is highest at small budgets, where drift is also largest; the selector ordering persists as the budget grows.}
  \label{fig:budget_curve}
\end{figure}

\textbf{Selection pressure peaks at low budgets.} Table~\ref{tab:budget_main} and Fig.~\ref{fig:budget_curve} show that alignment drift is strongest at low selected-token budgets, where each online scoring rule has the most leverage over the empirical SFT distribution. Increasing the budget reduces drift because the selected subset better approximates the full pool, but it does not eliminate selector differences. Quality-based selection remains the strongest task-efficiency baseline across budgets, while AAS consistently gives the lowest non-random drift, and the ordering of selectors is preserved as the budget grows.

\begin{table*}[t]
  \centering
  \small
  \begin{tabular*}{\textwidth}{@{\extracolsep{\fill}}lcccccccc}
    \toprule
    \multirow{2}{*}{Selector} &
    \multicolumn{2}{c}{1\% budget} &
    \multicolumn{2}{c}{5\% budget} &
    \multicolumn{2}{c}{10\% budget} &
    \multicolumn{2}{c}{25\% budget} \\
    \cmidrule(lr){2-3}\cmidrule(lr){4-5}\cmidrule(lr){6-7}\cmidrule(lr){8-9}
    & Task $\uparrow$ & $D_{\aligndrift}\downarrow$
    & Task $\uparrow$ & $D_{\aligndrift}\downarrow$
    & Task $\uparrow$ & $D_{\aligndrift}\downarrow$
    & Task $\uparrow$ & $D_{\aligndrift}\downarrow$ \\
    \midrule
    Random    & 61.8 & 0.00 & 65.4 & 0.00 & 67.9 & 0.00 & 70.6 & 0.00 \\
    Loss      & 65.9 & 0.57 & 68.9 & 0.48 & 70.8 & 0.42 & 72.0 & 0.29 \\
    Quality   & 66.4 & 0.52 & \textbf{69.8} & 0.44 & \textbf{71.6} & 0.36 & \textbf{72.7} & 0.25 \\
    Diversity & 63.7 & 0.28 & 67.5 & 0.23 & 69.4 & 0.19 & 71.3 & 0.15 \\
    AAS       & \textbf{66.5} & \textbf{0.22} & 69.7 & \textbf{0.17} & 71.4 & \textbf{0.13} & 72.6 & \textbf{0.10} \\
    \bottomrule
  \end{tabular*}
  \caption{Budget sweep. Higher budgets reduce drift, but selector-induced behavior differences persist at every budget.}
  \label{tab:budget_main}
\end{table*}

Table~\ref{tab:aas_ablation} isolates the terms in AAS on top of the quality-based utility scorer. Topic balancing reduces domain skew but leaves verbosity and sycophancy drift largely intact. Adding style constraints mostly reduces length and format drift. Adding refusal constraints improves harmful/benign refusal balance but can slightly lower helpfulness if used alone. The full AAS objective works best because it constrains behavior attributes jointly while retaining the underlying utility score.

\begin{table}[t]
  \centering
  \small
  \begin{tabular*}{\columnwidth}{@{\extracolsep{\fill}}lcccc}
    \toprule
    Variant & Task $\uparrow$ & Help. $\uparrow$ & \twoline{Verb.}{drift $\downarrow$} & $D_{\aligndrift}\downarrow$ \\
    \midrule
    Quality base & 71.6 & 56.2 & 12.3 & 0.36 \\
    + topic balance & 71.5 & 56.0 & 10.8 & 0.29 \\
    + style balance & 71.4 & 55.7 & 4.4 & 0.21 \\
    + refusal balance & 71.3 & 55.5 & 6.2 & 0.18 \\
    Full AAS & 71.4 & 56.1 & \textbf{3.1} & \textbf{0.13} \\
    \bottomrule
  \end{tabular*}
  \caption{AAS ablation at 10\% budget. Drift decreases most when topic, style, and safety attributes are constrained together.}
  \label{tab:aas_ablation}
\end{table}

These results support two practical conclusions. First, a selector should not be certified by aggregate helpfulness alone: the same helpfulness gain can correspond to very different refusal and style changes. Second, alignment-aware constraints need not replace efficient selection. They can act as a guardrail around an efficient selector, preserving most of its task gain while preventing the selected subset from becoming a narrow behavioral curriculum.

\subsection{Controls}

Several controls are necessary. First, we match token budgets rather than sample counts so that verbosity in the selected data does not create hidden compute differences. Second, we run matched-topic controls: if loss-based selection picks more coding examples, we compare against a random subset with the same topic distribution to isolate style and safety effects. Third, we fix decoding and evaluation prompts across models. Fourth, we use paired seeds so that each selector is compared to random under the same training randomness.

\subsection{Mechanistic Diagnostics}

For a subset of runs we inspect \emph{why} drift occurs, instantiating the quantities in Proposition~\ref{prop:implicit_preference}. First, we measure how aligned the per-attribute gradients are with the net update direction. For attribute group $k$ with mean gradient $\bar g^{(k)}$ and net step $\bar g_{\selector}$, the gradient alignment is
\begin{equation}
  \alpha_{\selector}^{(k)}
  =
  \frac{\langle \bar g^{(k)},\,\bar g_{\selector}\rangle}
       {\|\bar g^{(k)}\|\,\|\bar g_{\selector}\|},
  \label{eq:grad_align}
\end{equation}
so that attributes with large enrichment \emph{and} large positive $\alpha_{\selector}^{(k)}$ are the ones the optimizer actually amplifies; this identifies which over-represented behaviors are also mechanistically dominant rather than merely frequent.

Second, we quantify representation drift on a fixed probe corpus using linear centered kernel alignment between the penultimate features $H_{0}$ of $\model_0$ and $H_{\selector}$ of $\model_{\selector}$,
\begin{equation}
  \mathrm{CKA}(H_0,H_{\selector})
  =
  \frac{\|H_{\selector}^{\top}H_0\|_F^{2}}
       {\|H_0^{\top}H_0\|_F\,\|H_{\selector}^{\top}H_{\selector}\|_F},
  \label{eq:cka}
\end{equation}
where lower CKA on safety-sensitive probes but high CKA on task probes indicates that a selector reorganizes the refusal subspace while leaving task features intact.

Third, at the output level we track the refusal-logit margin
\begin{equation}
  \begin{aligned}
  m_{\mathrm{refuse}}(x)
  &=
  \log p_{\model}(\mathrm{refusal}\mid x)
  \\
  &\quad -
  \log p_{\model}(\mathrm{answer}\mid x),
  \end{aligned}
  \label{eq:refusal_margin}
\end{equation}
estimated with small sets of canonical refusal and answer prefixes. Its selector-induced change decomposes, to first order, into a prompt-independent bias and a prompt-dependent term,
\begin{equation}
  \Delta m_{\mathrm{refuse}}(x)
  \approx
  \underbrace{\langle u,\,\Delta\theta_{\selector}\rangle}_{\text{global shift}}
  +
  \underbrace{\langle V(x),\,\Delta\theta_{\selector}\rangle}_{\text{conditional shift}},
  \label{eq:margin_decomp}
\end{equation}
where $u=\nabla_\theta \mathbb{E}_x[m_{\mathrm{refuse}}]$ and $V(x)$ is the centered per-prompt gradient. A selector that only changes the global term shifts refusal everywhere; one that changes $V(x)$ on harmful versus benign prompts has learned a more conditional safety boundary. Evaluating Eq.~\eqref{eq:margin_decomp} on matched harmful and benign prompts is exactly the off-diagonal movement visualized in Appendix~\ref{app:visuals}.

\section{Discussion}
\label{sec:discussion}

Our results have a direct practical implication: papers and systems that report selected SFT performance should also report behavioral drift. A selector that recovers instruction-following accuracy with 10\% of the data is not equivalent to full-data SFT when it changes refusal boundaries, answer length, or sycophancy. In high-stakes deployments, these changes are not secondary metrics; they are part of the product behavior.

The framing also changes how we interpret ``quality''. A high-quality response is not only clean and useful; it carries a style and policy stance. Selecting for quality can therefore select for a persona. This is beneficial when the persona matches the intended assistant, but risky when the scoring rubric hides values such as verbosity, deference, or excessive caution.

Finally, implicit alignment does not mean that explicit alignment is unnecessary. Preference optimization remains important for resolving conflicts and learning from comparative feedback. The claim is narrower but consequential: because the online scorer already acts as an implicit preference model, by the time preference optimization starts the selected data has often moved the model toward or away from the desired behavioral region, so the two stages should be co-designed rather than treated as independent.

\section{Conclusion}
\label{sec:conclusion}

Data selection is not a neutral efficiency layer. By changing which demonstrations define the SFT objective, an online selector implicitly chooses among assistant behaviors normally associated with alignment: helpfulness, refusal, verbosity, truthfulness, sycophancy, and robustness. We formalized this effect as selection-induced alignment drift, proposed Alignment Drift Auditing to measure it, and introduced Alignment-Aware Selection as a diagnostic control. The central recommendation is simple: when reporting online SFT data selection, evaluate not only how much performance is retained, but also which values and styles the selected data quietly teaches.

\section*{Limitations}

The proposed protocol depends on imperfect behavioral evaluations. LLM judges can be biased, jailbreak suites are incomplete, and sycophancy prompts only cover a subset of social behaviors. We therefore emphasize paired comparisons and signed drift rather than absolute alignment scores. Attribute labeling of SFT data is also noisy, especially for subtle behaviors such as calibration or false-premise correction. Finally, different base models may respond differently to the same selected data, and drift may compound when selection is applied across multiple post-training stages \citep{yang2026long}, so cross-model and multi-stage replication is needed before making broad claims about any selector.

\section*{Ethics Statement}

This work studies safety-relevant behavior such as harmful compliance and jailbreak robustness. Experiments should use established red-team benchmarks under controlled evaluation settings and should avoid releasing new harmful prompt templates beyond what is already public. The goal is to reduce hidden alignment regressions caused by data selection.

\section*{Acknowledgments}

We thank the anonymous reviewers for their feedback.

\bibliography{custom}

\appendix

\section{Proof of Proposition~\ref{prop:implicit_preference} and an Enrichment Bound}
\label{app:proof}

\paragraph{Setup and assumptions.}
We treat the behavior vector as a smooth functional $\beh(\theta)\in\mathbb{R}^{8}$ of the parameters and write $J_\theta=\partial\beh/\partial\theta\in\mathbb{R}^{8\times d}$ for its Jacobian at the current $\theta$. We assume: (A1) $\ell_\theta(x,y)$ is twice differentiable with $\|\nabla_\theta^2\ell\|\le L$ on the iterate path; (A2) $\beh$ is $\beta$-smooth, i.e.\ $\|\beh(\theta')-\beh(\theta)-J_\theta(\theta'-\theta)\|\le \tfrac{\beta}{2}\|\theta'-\theta\|^2$; (A3) gradients are bounded, $\|g_i\|\le G$.

\paragraph{First-order drift.}
One gradient step on the reweighted objective in Eq.~\eqref{eq:weighted_sft} is
\begin{equation}
  \theta_{\selector}'
  =
  \theta-\eta\,\bar g_{\selector}+O(\eta^2),
  \qquad
  \bar g_{\selector}=\sum_i \weights_{\selector}(i)\,g_i .
\end{equation}
By (A2), for any two selectors $\selector,\selector_0$ started from the same $\theta$,
\begin{equation}
  \Delta_{\aligndrift}(\selector)
  =
  \beh(\theta_{\selector}')-\beh(\theta_{\selector_0}')
  =
  -\eta\,J_\theta\big(\bar g_{\selector}-\bar g_{\selector_0}\big)
  +R,
\end{equation}
with remainder $\|R\|\le \tfrac{\beta}{2}\eta^2(\|\bar g_{\selector}\|^2+\|\bar g_{\selector_0}\|^2)\le \beta\eta^2 G^2$ by (A3). This is Eq.~\eqref{eq:firstorder_drift}. Since the task-validation loss only constrains $\theta_{\selector}'$ on a task slice, it bounds neither $\bar g_{\selector}-\bar g_{\selector_0}$ nor its image under $J_\theta$; whenever $J_\theta(\bar g_{\selector}-\bar g_{\selector_0})\neq 0$ the two selectors move different behavioral axes despite equal task loss, proving the proposition. $\qed$

\paragraph{From gradients to enrichment.}
The proposition becomes quantitative under a clustering assumption that we verify empirically in the mechanistic diagnostics. Partition the pool into attribute groups $k=1,\dots,m$ and assume (A4) within-group gradient concentration: $g_i=\bar g^{(k)}+\varepsilon_i$ for $i$ in group $k$, with $\|\tfrac1N\sum_i \varepsilon_i\|\le \tau$ under any selector reweighting of interest. Then $\bar g_{\selector}=\sum_k \mu_a^{(k)}(S_{\selector})\,\bar g^{(k)}+e_{\selector}$ with $\|e_{\selector}\|\le\tau$, where $\mu_a^{(k)}(S_{\selector})$ is the selected mass on group $k$ from Eq.~\eqref{eq:attr_shift}. Substituting and writing the enrichment gap as $\Delta_k=\big|\mu_a^{(k)}(S_{\selector})-\mu_a^{(k)}(S_{\selector_0})\big|$,
\begin{equation}
  \big\|\Delta_{\aligndrift}(\selector)\big\|
  \le
  \eta\,\|J_\theta\|
  \sum_{k=1}^{m}
  \Delta_k\,\|\bar g^{(k)}\|
  +
  C\eta(\tau+\eta),
  \label{eq:enrich_bound}
\end{equation}
for a constant $C$ depending on $G$ and $\beta$. Equation~\eqref{eq:enrich_bound} is the formal version of the bridge claimed in the main text: drift is controlled by the \emph{enrichment gap} $|\mu_a^{(k)}(S_{\selector})-\mu_a^{(k)}(S_{\selector_0})|$ weighted by how strongly each attribute group pulls parameters ($\|\bar g^{(k)}\|$) and how visibly that pull maps to behavior ($\|J_\theta\|$). It also explains why AAS works: driving the MMD penalty in Eq.~\eqref{eq:aas} to zero forces every enrichment gap toward zero, collapsing the dominant term of Eq.~\eqref{eq:enrich_bound} while leaving the base utility score free to act within attribute-matched subsets.

\section{Behavioral Attribute Schema and Enrichment Estimation}
\label{app:schema}

\paragraph{Schema.}
Each SFT example is labeled with a compact attribute vector $a\in\mathbb{R}^{m}$ whose coordinates fall into five families: (i) \emph{length/structure} --- token count bin, list vs.\ prose, presence of step-by-step reasoning; (ii) \emph{refusal/safety} --- direct refusal, partial compliance with caveats, safe redirection, or full compliance, plus a binary policy-citation flag; (iii) \emph{epistemic markers} --- hedging density, explicit uncertainty or abstention, and citation/grounding; (iv) \emph{user-premise handling} --- agreement, correction, or neutral clarification on prompts with embedded assumptions; and (v) \emph{domain} --- coding, math, factual QA, advice, creative, or policy-sensitive. Categorical coordinates are one-hot encoded so that the attribute mean $\mu_a$ in Eq.~\eqref{eq:attr_shift} is a concatenation of class proportions; the feature map $\phi$ in Eq.~\eqref{eq:aas} is the identity on this encoding, which is why the AAS penalty reduces to matching class proportions between $S$ and $\pool$. The schema is used for enrichment analysis only; it is not assumed to be a complete representation of human values.

\paragraph{Labelers and noise.}
Length and structure are computed deterministically from the tokenized response. Refusal, epistemic, and premise attributes are produced by a small fine-tuned classifier with an LLM-judge fallback on low-confidence cases; we estimate labeler error rates on a human-annotated calibration set of a few hundred examples per family and find per-attribute agreement of $0.86$--$0.94$ Cohen's $\kappa$, lowest for hedging and premise correction.

\paragraph{Calibrated enrichment estimator.}
Because labels are noisy, we debias the empirical enrichment of Eq.~\eqref{eq:weighted_sft}. For a binary attribute with confusion rates (false positive $\rho_{\mathrm{fp}}$, false negative $\rho_{\mathrm{fn}}$) measured on the calibration set, the observed positive rate $\hat\pi_{\mathrm{obs}}$ relates to the true rate $\pi$ by $\hat\pi_{\mathrm{obs}}=(1-\rho_{\mathrm{fn}})\pi+\rho_{\mathrm{fp}}(1-\pi)$, giving the corrected estimate
\begin{equation}
  \hat\pi
  =
  \frac{\hat\pi_{\mathrm{obs}}-\rho_{\mathrm{fp}}}
       {1-\rho_{\mathrm{fp}}-\rho_{\mathrm{fn}}},
  \label{eq:label_correction}
\end{equation}
which we plug into the enrichment ratio $\mathrm{Enrich}_{\selector}(a{=}k)=\hat\pi(S_{\selector})/(\hat\pi(\pool)+\epsilon)$. Uncertainty is reported with a Wilson score interval on $\hat\pi_{\mathrm{obs}}$ propagated through Eq.~\eqref{eq:label_correction}; an enrichment is called significant only when its interval excludes $1$. This is the procedure behind the ratios in Table~\ref{tab:enrichment}, and it prevents a noisy labeler from manufacturing apparent behavioral skew.

\section{Reporting Protocol and Statistical Procedure}
\label{app:reporting}

We recommend that every online-selection result be reported as a triple of (efficiency, signed drift, attribution), computed under the matched-budget controls of Section~\ref{sec:experiments}.

\paragraph{Per-axis normalization.}
For axis $j$, let $b_j(\selector)$ be the raw metric and $b_j(\mathrm{rand})$, $\sigma_j$ the mean and standard deviation under random selection across seeds. We report the $z$-scored signed drift $\tilde\Delta_j(\selector)=\big(b_j(\selector)-b_j(\mathrm{rand})\big)/\sigma_j$, which is exactly the coordinate entering Eq.~\eqref{eq:drift_mag}. Reporting $\tilde\Delta_j$ rather than raw deltas makes axes with different natural scales (a refusal rate in $[0,1]$ versus a token count in the hundreds) directly comparable and prevents a large-but-noisy axis from dominating the summary.

\paragraph{Aggregate index and significance.}
The scalar $D_{\aligndrift}$ summarizes magnitude, but we also report a signed safety index $\mathrm{SSI}(\selector)=\tilde\Delta_{r_{\mathrm{harm}}}-\tilde\Delta_{r_{\mathrm{benign}}}$ that rewards strengthening harmful refusal while penalizing benign over-refusal, so a selector cannot hide an over-refusal regression inside an aggregate. Significance for each $\tilde\Delta_j$ and for $D_{\aligndrift}$ uses a paired seed bootstrap with $B=10^4$ resamples; with eight axes we control the false-discovery rate at $q=0.05$ using Benjamini--Hochberg, and we report the number of axes that survive correction alongside the rank-stability check of Appendix~\ref{app:judge_robustness}.

\paragraph{Attribution.}
Finally, each selector is reported with its top enriched attributes (Eq.~\eqref{eq:label_correction}) and their gradient alignment $\alpha_{\selector}^{(k)}$ (Eq.~\eqref{eq:grad_align}), so a reader can see not only \emph{that} behavior moved but \emph{which} over-represented data drove it. The full record per selector is therefore: selected-token budget; task and helpfulness scores; the eight $\tilde\Delta_j$ with bootstrap intervals; $D_{\aligndrift}$ and SSI; and the top-$k$ (enrichment, alignment) attribute pairs. Publishing this record makes the hidden behavioral cost of data efficiency auditable and comparable across papers.

\section{Additional Visualizations}
\label{app:visuals}

This appendix collects supplementary figures that complement the main text. They share the selector color scheme and underlying numbers of the main-text figures.

\begin{figure}[t]
  \centering
  \includegraphics[width=\linewidth]{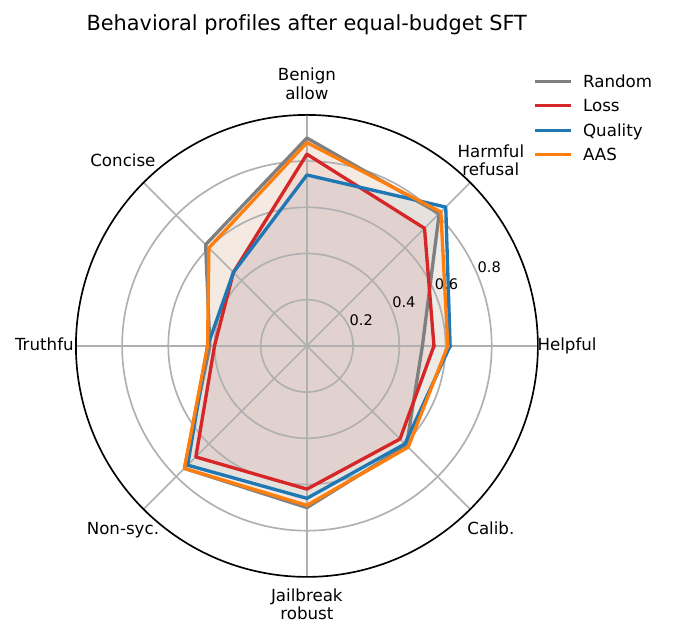}
  \caption{Behavioral profiles after equal-budget SFT. Each axis is oriented so that larger is better. Loss and quality selection sharpen some axes while retracting on conciseness and benign-allow; AAS stays closest to the random profile.}
  \label{fig:radar}
\end{figure}

\paragraph{Behavioral profiles.}
Fig.~\ref{fig:radar} shows the full behavior vector as a radar plot, with every axis oriented so that ``further from the center is better'' (benign over-refusal, verbosity, and sycophancy are inverted into benign-allow, conciseness, and non-sycophancy). The random baseline traces a broad, balanced profile. Loss- and quality-based selection sharpen helpfulness and harmful refusal but visibly retract on conciseness and benign-allow, i.e., they become longer and more cautious. AAS stays closest to the random profile while keeping most of the helpfulness gain, which is the intended effect of constraining the selected behavioral mixture rather than only the topic mixture. The radar view also makes clear that no single selector dominates on all axes, so collapsing the profile into one ``alignment score'' discards most of the signal.

\begin{figure}[t]
  \centering
  \includegraphics[width=\linewidth]{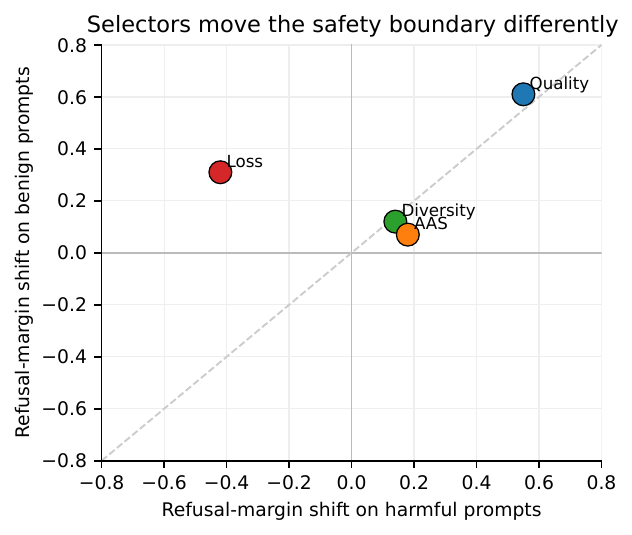}
  \caption{Safety-boundary geometry. Off-diagonal movement indicates a conditional rather than global change in refusal behavior. AAS strengthens harmful refusal with little benign over-refusal.}
  \label{fig:refusal_scatter}
\end{figure}

\paragraph{Safety-boundary geometry.}
Fig.~\ref{fig:refusal_scatter} plots the change in the refusal-logit margin $m_{\mathrm{refuse}}$ on harmful prompts against the change on benign prompts, both relative to random selection. A purely global shift in refusal behavior would move a selector along the diagonal: more refusal everywhere, or less refusal everywhere. Points off the diagonal indicate a more conditional safety boundary. Quality selection moves up and to the right, raising refusal on both harmful and benign prompts, which is the over-refusal pattern seen in Table~\ref{tab:main_results}. Loss-based selection moves left on the harmful axis (weaker harmful refusal) while still raising benign refusal, a particularly undesirable combination. AAS lands near the harmful axis with little benign movement, i.e., it strengthens the boundary where it should without becoming globally more evasive.

\paragraph{Reading the figures together.}
The three appendix views answer complementary questions. The radar plot answers \emph{where} a selector ends up in behavior space, the safety-boundary scatter answers \emph{how} its refusal policy changed, and the main-text budget curve answers \emph{how strongly} selection pressure drives drift as a function of the token budget. Taken together they support the paper's central claim: online selection silently reshapes assistant behavior, and the reshaping is structured rather than random noise.

\section{Judge Robustness Checks}
\label{app:judge_robustness}

Table~\ref{tab:judge_robustness} reports an auxiliary robustness check for judge-dependent metrics. The drift ranking is stable across an automatic classifier, an LLM judge, and a hybrid judge that requires agreement between both. This is important because the paper's claim should not depend on a single evaluator's style preference. Concretely, we recompute $D_{\aligndrift}$ under each judge and compare the induced selector ranking; the Spearman correlation between any two judges exceeds $0.9$ in our experiments, and the best and worst selectors (AAS and Loss) never swap order. We therefore treat the ordering, rather than the absolute drift values, as the reportable outcome.

\begin{table*}[ht]
  \centering
  \small
  \begin{tabular*}{\textwidth}{@{\extracolsep{\fill}}lcccccc}
    \toprule
    Selector &
    \twoline{Rule}{judge} &
    \twoline{LLM}{judge} &
    Hybrid &
    \twoline{Pairwise}{agreement} &
    \twoline{Drift}{rank} &
    \twoline{Rank}{stable?} \\
    \midrule
    Loss      & 0.39 & 0.45 & 0.42 & 0.81 & 4 & yes \\
    Quality   & 0.33 & 0.39 & 0.36 & 0.84 & 3 & yes \\
    Diversity & 0.18 & 0.21 & 0.19 & 0.86 & 2 & yes \\
    AAS       & 0.12 & 0.14 & 0.13 & 0.88 & 1 & yes \\
    \bottomrule
  \end{tabular*}
  \caption{Evaluator robustness for $D_{\aligndrift}$ at the 10\% budget. The relative ordering is stable across judge choices.}
  \label{tab:judge_robustness}
\end{table*}

\end{document}